\documentclass[11pt,a4paper]{article}
\usepackage[hyperref]{acl2019}
\usepackage{times}
\usepackage{latexsym}
\usepackage{graphicx}
\usepackage{url}
\usepackage{multirow}
\usepackage{hyperref}
\usepackage{cleveref}
\aclfinalcopy 

\title{A Deep Learning Approach for Automatic Detection of Fake News}
\author{
Tanik Saikh$^1$ \quad Arkadipta De$^2$ \quad Asif Ekbal$^{1}$ \quad Pushpak Bhattacharyya$^{1}$\\
Department of Computer Science and Engineering$^1$ \\
Indian Institute of Technology Patna$^1$ \\
Department of Computer Science and Engineering$^2$ \\
Government College of Engineering and Textile Technology, Berhampore$^2$ \\ 
\texttt{\{tanik.srf17, asif, pb\}@iitp.ac.in$^1$} \\
\texttt{de.arkadipta05@gmail.com$^2$} \\
}

\date{}

\begin{document}
\maketitle
\begin{abstract}
Fake news detection is a very prominent and essential task in the field of journalism. This challenging problem is seen so far in the field of politics, but it could be even more challenging when it is to be determined in the multi-domain platform. In this paper, we propose two effective models based on deep learning for solving fake news detection problem in online news contents of multiple domains. We evaluate our techniques on the two recently released datasets, namely \textit{FakeNews AMT and Celebrity} for fake news detection. The proposed systems yield encouraging performance, outperforming the current handcrafted feature engineering based state-of-the-art system with a significant margin of 3.08\% and 9.3\% by the two models, respectively. In order to exploit the datasets, available for the related tasks, we perform cross-domain analysis (i.e. model trained on FakeNews AMT and tested on Celebrity and vice versa) to explore the applicability of our systems across the domains.
\end{abstract}



\section{Introduction}
In the emergence of social and news media, data are constantly being created day by day. The data so generated are enormous in amount, and often contains miss-information. Hence it is necessary to check it's truthfulness. Nowadays people mostly rely on social media and many other online news feeds as their only platforms for news consumption \cite{social-media}. A survey from the \textit{Consumer News and Business Channel (CNBC)} also reveals that more people are rely on social media for news consumption rather than news paper \footnote{https://www.cnbc.com/2018/12/10/social-media-more-popular-than-newspapers-for-news-pew.html}. Therefore, in order to deliver the genuine news to such consumers, checking the truthfulness of such online news content is of utmost priority to news industries. The task is very difficult for a machine as even human being can not understand news article's veracity (easily) after reading the article. \\
Prior works on fake news detection entirely rely on the datasets having satirical news contents sources, namely \textit{"The Onion" \cite{rubin2016fake}}, fact checking website like \textit{Politi-Fact \cite{wang-2017-liar}, and “Snopes” \cite{DBLP:conf/cikm/PopatMSW16}}, and on the contents of the websites which track viral news such as \textit{BuzzFeed \cite{potthast-etal-2018-stylometric}} etc. But these sources have severe drawbacks and multiple challenges too. Satirical news mimic the real news which are having the mixture of irony and absurdity. Most of the works in fake news detection fall in this line and confine in one domain (i.e. politics). The task could be even more challenging and generic if we study this fake news detection problem in multiple domain scenarios. We  endeavour to mitigate this particular problem of fake news detection in multiple domains. This task is even more challenging compared to the situation when news is taken only from a particular domain, i.e. uni-domain platform. We make use of the dataset which contained news contents from multiple domains. 
The problem definition would be as follows: \\
Given a \textit{News Topic} along with the corresponding \textit{News Body Document}, the task is to classify whether the given news is \textit{legitimate/genuine} or \textit{Fake}.
The work described in \citet{C18-1287} followed this path. They also offered two novel computational resources, namely \textit{FakeNews AMT} and \textit{Celebrity news}. These datasets are having triples of \textit{topic, document and label (Legit/Fake)} from multiple domains (like \textit{Business, Education, Technology, Entertainment and Sports} etc) including \textit{politics}. Also, they claimed that these datasets focus on the deceptive properties of online articles from different domains. They provided a baseline model. The model is based on Support Vector Machine (SVM) that exploits the hand-crafted linguistics features. The SVM based model achieved the accuracies of 74\% and 76\% in the FakeNews AMT and Celebrity news datasets, respectively. We pose this problem as a classification problem. So the proposed predictive models are binary classification systems which aim to classify between fake and the verified content of online news from multiple domains. We solve the problem of multi-domain fake news detection using two variations of deep learning approaches. The first model (denoted as Model 1) is a Bi-directional Gated Recurrent Unit (BiGRU) based deep neural network model, whereas the second model (i.e. Model 2) is \textit{Embedding from Language Model (ELMo)} based. It is to be noted that the use of deep learning to solve this problem in this particular setting is, in itself, very new. The technique, particularly the word attention mechanism, has not been tried for solving such a problem. Existing prior works for this problem mostly employ the methods that make use of handcrafted features. The proposed systems do not depend on hand crafted feature engineering or a sophisticated NLP pipeline, rather it is an end to end deep neural network architecture. Both the models outperform the state-of-the-art system.
\section{Related Work}
 A sufficient number of works could be found in the literature in fake news detection. Nowadays the detection of fake news  is a hot area of research and gained much more research interest among the researchers. We could detect fake news at two levels, namely the conceptual level and  operational level. \citet{Rubin:2015:DDN:2857070.2857153} defined that conceptually there are three types of fake news: \textit{viz i. Serious Fabrications ii. Hoaxes and iii. Satire}. The work of  \citet{conroy2015automatic} fostered linguistics and fact checking based approaches to distinguish between real and fake news, which could be considered as the work at conceptual level. \citet{chen2015news} described that fact-checking approach is a verification of hypothesis made in a news article to judge the truthfulness of a claim. \citet{thorne-etal-2018-fever} introduced a novel dataset for fact-checking and verification where evidence is large Wikipedia corpus. Few notable works which made use of text as evidence can be found in  \cite{ferreira-vlachos-2016-emergent,nie2018combining}. \\
 The Fake News Challenge \footnote{http://www.fakenewschallenge.org/} organized a competition to explore, how artificial intelligence technologies could be fostered to combat fake news. Almost 50 participants were participated and submitted their systems. \citet{hanselowski-etal-2018-retrospective} performed retrospective analysis of the three best participating systems of the Fake News Challenge. The work of \citet{saikh2019novel} detected fake news through stance detection and also correlated this stance classification problem with Textual Entailment (TE). They tackled this problem using statistical machine learning and deep learning approaches separately and with combination of both of these. This system achieved the state of the art result. \\
 Another remarkable work in this line is the verification of a human- generated claim given the whole Wikipedia as evidence. The dataset, namely (\textit{Fact Extraction and Verification (FEVER)}) proposed by \citet{thorne-etal-2018-fever} served this purpose. Few notable works in this line could be found in \cite{yin-roth-2018-twowingos,DBLP:conf/aaai/NieCB19}.   
 \section{Proposed Methods}
We propose two deep Learning based models to address the problem of fake information detection in the multi-domain platform. In the following subsections, we will discuss the methods.
\subsection{Model 1}
\begin{figure*}
\centering
  \includegraphics[height=3cm, width=15cm]{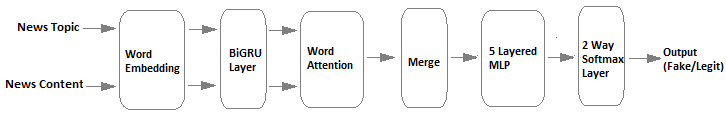}
  \caption{Architectural Diagram of the Proposed First System}
  \label{fig1}
\end{figure*}

\begin{figure}
\centering
  \includegraphics[height = 5cm, width = 8cm]{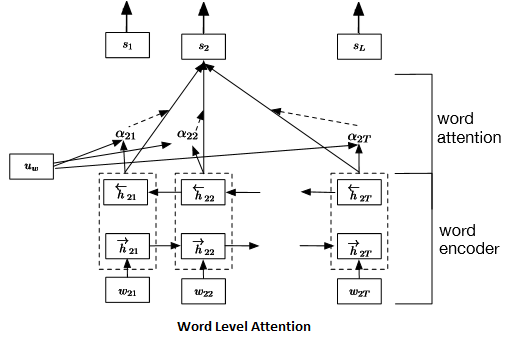}
  \caption{Word Level Attention Network}
  \label{fig-attention}
\end{figure}

This model comprises of multiple layers as shown in the Figure \ref{fig1}. The layers are \textit{A. Embedding Layer B. Encoding Layer (Bi-GRU) C. Word level Attention D. Multi-layer Perceptron (MLP)}.\\ 
\textit{\textbf{A. Embedding Layer:}} The embedding of each word is obtained using pre-trained fastText model\footnote{https://fasttext.cc/}\cite{bojanowski2017enriching}. FastText embedding model is an extended version of Word2Vec \cite{NIPS2013_5021}. Word2Vec (predicts embedding of a word based on given context and vice-versa) and Glove (exploits count and word co-occurrence matrix to predict embedding of a word) \cite{pennington-etal-2014-glove} both
treat each word as an atomic entity. The fastText model produces embedding of each word by combining the embedding of each character n-gram of that word. The model works better on rare words and also produces embedding for out-of-vocabulary words, where Word2Vec and Golve both fail. In the multi-domain scenario vocabularies are from different domains and there is a high chance of existing different domain specific vocabularies. This is the reason for choosing the fastText word vector method.\\ 
\textit{B. \textbf{Encoding Layer:}} The representation of each word is further given to a bidirectional Gated Recurrent Units (GRUs) \cite{DBLP:journals/corr/ChoMGBSB14} model. GRU takes less parameter and resources compared to Long Short Term Memory (LSTM), training also is computationally efficient. The working principles of GRU obey the following equations: \\

\begin{equation}\label{eq1}
z = \alpha (x_tU^z+s_{t-1}W^z)
\end{equation}
\begin{equation}\label{eq2}
r = \alpha (x_tU^r+s_{t-1}W^r)
\end{equation}
\begin{equation}\label{eq3}
h = tanh (x_tU^h+r_t \cdot s_{t-1} W^r)
\end{equation}
\begin{equation}\label{eq4}
r = (1-z)\cdot h+ z\cdot s_{t-1}
\end{equation}
In equation \ref{eq1}, z is the update gate at time step t. This z is the summation of the multiplications of $x_t$ with it's own weight U(z) and $s_{t-1}$ (holds the information of previous state) with  it's own W(z). A sigmoid $\alpha$ is applied on the summation to squeeze the result between 0 and 1. The task of this update gate (z) is to help the model to estimate how much of the previous information (from previous time steps) needs to be passed along to the future. In the equation \ref{eq2}, r is the reset gate, which is responsible for taking the decision of how much past information to forget. The calculation is same as the equation \ref{eq1}. The differences are in the weight and gate usages. The equation \ref{eq3} performs as follows, i. multiply input $x_{t}$ with a weight U and $s_{t-1}$ with a weight W. ii. Compute the element wise product between reset gate $r_t$ and $s_{t-1}$W. Then a non-linear activation function \textit{tanh} is applied to the summation of i and ii. 
Finally, in the equation \ref{eq4}, we compute r which holds the information of the current unit. The computation procedure is as follows: i. compute element-wise multiplication to the update gate $z_t$ and $s_{(t-1)}$. ii. calculate element-wise multiplication to (1-z) with h. Take the summation of i and ii.\\
The bidirectional GRUs consists of the forward GRU, which reads the sentence from the first word ($w_1$) to the last word ($w_L$) and the backward GRU, that reads in reverse direction. We concatenate the representation of each word obtained from both the passes. \\
\textit{C. \textbf{Word Level Attention:}} We apply the attention model at word level \cite{bahdanau2015neural,xu2015show}. The objective is to let the model decide which words are importance compared to other words while predicting the target class (fake/legit). We apply this as applied in \citet{yang2016hierarchical}. The diagram is shown in the Figure \ref{fig-attention}. We take the aggregation of those words' representation which are multiplied with attention weight to get sentence representation. We do this process for both the news topic and the corresponding document. This particular technique of the word attention mechanism, has not been tried for solving such a problem. \\

 \begin{equation}\label{eq5}
 U_{i_t} = tanh(W_wh_{i_t}+b_w)
 \end{equation}
 \begin{equation}\label{eq6}
 \alpha_{i_t} = \frac{exp(u_{i_t}^Tu_w)}{\sum_{t}exp(u_{i_t}^Tu_w)}
 \end{equation}
 \begin{equation}\label{eq7}
 s_i = \sum_{t}\alpha_{i_t}h_{i_t}
 \end{equation}
 First get the word annotation $h_{i_t}$ through GRU output and compute $u_{i_t}$ as a hidden representation of $h_{i_t}$ in \ref{eq5}. We measure the importance of the word as the similarity of $u_{i_t}$ with a word level context vector $u_w$ and get a normalized importance weight $\alpha_{i_t}$ through a softmax in \ref{eq6}. After that, in \ref{eq7}, we compute the sentence vector $s_i$ as a weighted sum of the word annotations based on the weights $\alpha_{i_t}$. The word context vector $u_w$ is randomly initialized and jointly learned during the training process. \\ 
\textit{\textbf{D. Multi-Layer Perceptron:}} We concatenate the sentence vector obtained for both the inputs. The obtained vector further fed into fully connected layers. We use 512, 256, 128, 50 and 10 neurons, respectively, for five such layers with ReLU \cite{glorot2011deep} activation in each layer. Between each such layer, we employ 20\% dropout \cite{srivastava2014dropout} as a measurement of regularization. Finally, the output from the last fully connected layer is fed into a final classification layer with softmax \cite{Duan:2003:MCS:1764295.1764312} activation function having 2 neurons. We use Adam \cite{kingma2014adam} optimizer for optimization. 
\subsection{Model 2}
\begin{figure*}
\centering
  \includegraphics[height=3cm, width=15cm]{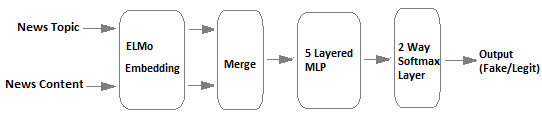}
  \caption{Architectural Diagram of the Proposed Second Model}
  \label{fig2}
\end{figure*}

We propose another approach whose embedding layer is based on \textit{Embedding for Language Model (ELMo)} \cite{N18-1202} and the MLP Network, which is same as we applied in Model 1. The diagram of this model is shown in the Figure \ref{fig2}.  

\textit{\textbf{Embedding Layer:}} Embedding from Language Model (ELMo) has several advantages over the other word vector methods, and found to be a good performer in many challenging NLP problems. It has key features like i. Contextual i.e. representation of each word is based on entire corpus in which it is used ii. Deep i.e. it combines all layers of a deep pre-trained neural network and iii. Character based i.e. it provides representations which are based on character, thus allowing the network to make use of morphological clues to form robust representation of out-of-vocabulary tokens during training. The ELMO embedding is very efficient in capturing context. The multi-domain datasets are having different vocabularies and contexts, so we make use of such a word vector representation method to capture the context. News topics and corresponding documents are given to Elmo Embedding model. This embedding layer produces the representation for news topic and news content.\\
After getting the embedding of the topic and the context, we merge them. The merged vector is fed into a five layers MLP (same as the previous model). Finally, we classify with a final layer having softmax activation function.
\section{Experiments, Results and Discussion and Comparison with State-of-the-Art}
Overall we perform four sets of experiments. In the following sub-sections we describe and analyze them one by one after the description of the datasets used.\\
\textit{\textbf{Data:}}
Prior datasets and focus of research for fake information detection are on political domain. 
As our research focus is on multiple domains, we foster the dataset released by \citet{C18-1287}. They released two novel datasets, namely \textit{FakeNews AMT} and \textit{Celebrity}. The Fake News AMT is collected via crowdsourcing (Amazon Mechanical Turk (AMT)) which covers news of six domains (i.e. Technology, Education, Business, Sports, Politics, and Entertainment). The Celebrity dataset is crawled directly from the web of celebrity gossips. It covers celebrity news. The AMT manually generated fake version of a news based on the real news. 
We extract the data domain wise to get the statistics of the dataset. It is observed that each domain contains equal number of instances (i.e. 80). The class distribution among each domain is also evenly distributed. The statistics of these two datasets is shown in the following Table \ref{data-stat}. \\

\begin{table}[]
\scriptsize
\centering
\begin{tabular}{|c|c|c|c|c|}
\hline
Dataset & \# of Examples & Avg.words/sent & Words & Label \\ \hline
\multirow{2}{*}{FakeNewsAMT} & 240 & 132/5 & 31,990 & Fake \\ \cline{2-5} 
 & 240 & 139/5 & 33,378 & Legit \\ \hline
\multirow{2}{*}{Celebrity} & 250 & 399/17 & 39,440 & Fake \\ \cline{2-5} 
 & 250 & 700/33 & 70,975 & Legit \\ \hline
\end{tabular}
\caption{Class Distribution and Word Statistics for Fake News AMT and Celebrity Datasets. Avg: Average, sent: Sentence}
\label{data-stat}
\end{table}
The news of the Fake News AMT dataset was obtained from a variety of mainstream news websites predominantly in the United States such as the ABCNews, CNN, USAToday, New York Times, FoxNews, Bloomberg, and CNET among others. 

\begin{figure}
\centering
  \includegraphics[height=0.6cm, width=7.5cm]{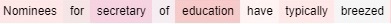}
  \caption{Word Level Attention on News Topic}
  \label{fig3}
\end{figure}

\begin{figure}
\centering
  \includegraphics[height=0.6cm, width=7.5cm]{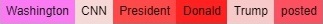}
  \caption{Word level Attention on News Document, A Part of it is Shown Due to Space Constraint.}
  \label{fig4}
\end{figure}

\textit{\textbf{Multi-Domain Analysis:}} In this section, we do experiments on whole Fake News AMT and Celebrity datasets individually. We train our models on the whole Fake News AMT and Celebrity dataset and test on the respective test set. As the datasets is evenly distributed between real and fake news item, a random baseline of 50\% could be assumed as reference. The results obtained by the two proposed methods outperform the baseline and the results of \citet{C18-1287}. The results obtained and comparisons are shown in the Table \ref{res-tab}. Our results indicate this task could be efficiently handled using deep learning approach.\\
\begin{table}[]
\scriptsize
\centering
\resizebox{\linewidth}{!}{
\begin{tabular}{|l|l|l|l|}
\hline
\textbf{Dataset} &  \textbf{System} & \textbf{Model} & \textbf{Test Accuracy(\%)} \\ \hline
\multirow{3}{*}{FakeNews AMT} & \multirow{2}{*}{Proposed} & Model1 & 77.08 \\ \cline{3-4} 
 &  & Model2 & 83.3 \\ \cline{2-4} 
 & \cite{C18-1287} & Linear SVM & 74 \\ \hline
\multirow{3}{*}{Celebrity} & \multirow{2}{*}{Proposed} & Model1 & 76.53 \\ \cline{3-4} 
 &  & Model2 & 79 \\ \cline{2-4} 
 & \cite{C18-1287} & Linear SVM & 76 \\ \hline
\end{tabular}}
\caption{Classification Results for the FakeNews AMT and Celebrity News Dataset with Two Proposed Methods and Comparison with Previous Results}
\label{res-tab}
\end{table}
\textit{\textbf{Cross-Domain Analyses:}} We perform another set of experiment to study the usefulness of the best performing system (i.e. Model2 ) across the domains. We train the model2, on FakeNews AMT and test on Celebrity and vice-versa. The results are shown in the Table \ref{CrossDomain}. If we compare with the in domain results it is observed that there is a significant drop. This drop also observed in the work of \citet{C18-1287} in machine learning setting. This indicates there is a significant role of a domain in fake news detection, as it is established by our deep learning guided experiments too.
\begin{table}[]
\scriptsize
\centering
\resizebox{\linewidth}{!}{
\begin{tabular}{|l|l|l|}
\hline
\textbf{Training} & \textbf{Testing} & \textbf{Accuracy(\%)} \\ \hline
FakeNewsAMT & Celebrity & 54.3 \\ \hline
Celebrity & FakeNewsAMT & 68.5 \\ \hline
\end{tabular}}
\caption{Results Obtained in Cross-Domain Analysis Experiments on the Best Performing System.}
\label{CrossDomain}
\end{table}
\\
\textit{\textbf{Multi-Domain Training and Domain-wise Testing:}} 
There are very small number of examples pairs in each sub-domain (i.e. Business, Technology etc) in FakeNews AMT dataset. We combine the examples pairs of multiple domains/genres for cross corpus utilization. We train our proposed models on the combined dataset of five out of six available domains and test on the remaining one. This has been performed to see how the model which is trained on heterogeneous data react on the domain to which the model was not exposed at the time of training. The results are shown in \textit{Exp. a} part of the Table \ref{resu-exp-a-b}. 
Both the models yield the best accuracy in the  \textit{Education} domain, which indicates this domain is open i.e. linguistics properties, vocabularies of this domain are quite similar to other domains. The models (i.e. Model 1 and 2) perform worst in the \textit{Entertainment} and the  \textit{Sports}, respectively, which indicate these two domains are diverse in nature from the others in terms of linguistics properties, writing style, vocabularies etc.\\
\textit{\textbf{Domain-wise Training and Domain-wise Testing:}}
We also eager to see in-domain effect of our systems. The FakeNews AMT dataset comprises of six separate domains. We train and test our models, on each domain's dataset of Fake News AMT. 
This evaluates our model's performance domain-wise. 
The results of this experiment are shown in the \textit{Exp. b} part of the Table \ref{resu-exp-a-b}. In this case both the models produce the highest accuracy in the \textit{Sports} domain, followed by the \textit{Entertainment}, as we have shown in our previous experiments that these two domains are diverse in nature from the others. This fact is established by this experiment too. Both the models produce the lowest result in the \textit{Technology} and the \textit{Business} domain, respectively.\\ 
\begin{table}[]
\scriptsize
\centering
\begin{tabular}{|c|c|c|l|l|}
\hline
\multirow{2}{*}{\textbf{Domain}} & \multicolumn{2}{c|}{\textbf{Exp. a}} & \multicolumn{2}{c|}{\textbf{Exp. b}} \\ \cline{2-5} 
 & \textbf{Model1} & \textbf{Model2} & \textbf{Model1} & \textbf{Model2} \\ \hline
Business & 74.75 & 78.75 & 63.56 & 68.56 \\ \hline
Education & 77.25 & 91.25 & 65.65 & 70.65 \\ \hline
Technology & 76.22 & 88.75 & 64.3 & 65.35 \\ \hline
Politics & 73.75 & 88.75 & 64.27 & 69.22 \\ \hline
Entertainment & 68.25 & 76.25 & 65.89 & 71.2 \\ \hline
Sports & 70.75 & 73.75 & 67.86 & 71.45 \\ \hline
\end{tabular}
\caption{Result of Exp. a (Trained on Multi-domain Data and Tested on Domain wise Data) and Exp. b (Trained on Domain wise Data and Tested on Domain wise Data)}
\label{resu-exp-a-b}
\end{table}
\textit{\textbf{Visualization of Word Level Attention:}} We take the visualization of the topic and the corresponding document at word level attention as shown in the Figure \ref{fig3} and \ref{fig4}, respectively. The aim is to visualize the words which are assigned more weights during the prediction of the output class. In these Figures, words with more deeper colour indicate that they are getting more attention. We can observe, the words \textit{secretary}, \textit{education} in \ref{fig3} and \textit{President}, \textit{Donald} in \ref{fig4} are the words having deeper colour, i.e. these words are getting more weight compared to others. These words are Named Entities (NEs). It could be concluded that NEs phrases are important in fake news detection in multi domain setting.
\subsection{Error Analysis}
We extract the mis-classified and also the truly classified instances produced by the best performing system. We perform a rigorous analysis of these instances and try to find out the pattern in the mis-classified instances and the linguistics differences between those two categories of instances. It is found that the model fails mostly in the \textit{Entertainment} followed by the \textit{sports} and the  \textit{Business} domain etc. To name a few, we are showing such examples which are actually "Legitimate", but predicted as "Fake" the Table \ref{tab:misclassified ins} and which are actually "Fake", but predicted as "Legitimate" the Table \ref{tab:misclassified ins-1}. It is observed that both the topic and document are having ample number of NEs. It needs further investigation in this font.
\begin{table*}[]
\centering
\scriptsize
\begin{tabular}{|l|l|l|}
\hline
\multicolumn{1}{|c|}{\textbf{Domain}} & \multicolumn{1}{c|}{\textbf{Topic}} & \multicolumn{1}{c|}{\textbf{Content}} \\ \hline
Entertainment & \begin{tabular}[c]{@{}l@{}}Chris Pratt responds to body \\ shamers telling him he's too thin\end{tabular} & \begin{tabular}[c]{@{}l@{}}Big or small Chris Pratt has heard it all. These days the \\ "Guardians of the Galaxy" star  37  is taking flak for \\ being too thin but he's not taking it lying down. Pratt  \\ who has been documenting the healthy snacks he's eating \\ while filming "Jurassic World 2" in a series of \\ "What's My Snack" Instagram videos  fired back -- in his \\ usual tongue-in-cheek manner -- after some followers  \\ apparently suggested he looked too thin. "So many \\ people have said I look too thin in my recent episodes of \\ \#WHATSMYSNACK  he wrote on Instagram Thursday.\\ Some have gone as far as to say I look 'skeletal.' \\ Well just because I am a male doesn't mean I'm\\ impervious to your whispers. Body shaming hurts."\end{tabular} \\ \hline
Business & \begin{tabular}[c]{@{}l@{}}Banks and Tech Firms Battle Over \\ Something Akin to Gold: Your Data\end{tabular} & \begin{tabular}[c]{@{}l@{}}The big banks and Silicon Valley are waging an escalating\\ battle over your personal financial data: your dinner bill last\\ night your monthly mortgage payment the interest rates you\\ pay. Technology companies like Mint and Betterment have \\ been eager to slurp up this data mainly by building services \\ that let people link all their various bank-account and \\ credit-card information. The selling point is to make \\ budgeting and bookkeeping easier. But the data is also \\ being used to offer new kinds of loans and investment\\ products. Now banks have decided they aren't letting \\ the data go without a fight. In recent weeks several \\ large banks have been pushing to restrict the sharing\\ of this kind of data with technology companies according \\ to the tech firms. In some cases they are refusing to pass \\ along information like the fees and interest rates \\ they charge. Both sides see big money to be made \\ from the reams of highly personal information created\\ by financial transactions.\end{tabular} \\ \hline
\end{tabular}
\caption{Examples of mis-classified instances from Entertainment and Business domain. Examples are originally "Legitimate" but predicted as "Fake".}
\label{tab:misclassified ins}
\end{table*}
\begin{table*}[]
\centering
\scriptsize
\begin{tabular}{|l|l|l|}
\hline
\multicolumn{1}{|c|}{\textbf{Domain}} & \multicolumn{1}{c|}{\textbf{Topic}} & \multicolumn{1}{c|}{\textbf{Content}} \\ \hline
Sports & \begin{tabular}[c]{@{}l@{}}Slaven Bilic still has no support of \\ West Ham's owners\end{tabular} & \begin{tabular}[c]{@{}l@{}}"West Ham's owners have no faith in manager Slaven Bilic \\ as his team won only six of their 11 games this year \\ according to Sky sources. Bilic's contract runs out in the \\ summer of 2018 and results have made it likely that he\\ will not be offered a new deal this summer. Co-chairman \\ David Sullivan told supporters 10 days ago after \\ West Ham  lost 3-2 at home to Leicester City. Sullivan said \\ that even if performances and results improved in the next \\ three games against Hull City Arsenal and Swansea City. \\ West Ham's owners have a track record of being unloyal \\ to their managers who don't meet their specs and there is\\ a acceptance at boardroom level that Bilic has failed to \\ prove a solid season."\end{tabular} \\ \hline
Education & \begin{tabular}[c]{@{}l@{}}STEM Students Create \\ Winning Invention\end{tabular} & \begin{tabular}[c]{@{}l@{}}STREAMWOOD, Ill. (AP) -- A group of Streamwood High \\ School students have created an invention that is \\ exciting homeowners everywhere - and worrying \\ electricity companies at the same time. The kids \\ competed in the Samsung Solve for Tomorrow\\ contest, entering and winning with a new solar panel \\ that costs about \$100 but can power an entire home \\ - no roof takeover needed!  The contest won the \\ state-level competition which encourages teachers \\ and students to solve real-world issues using science \\ and math skills; the 16 studens will now compete in a\\ national competition and, if successful, could win a \\ prize of up to \$200,000.\end{tabular} \\ \hline
\end{tabular}
\caption{Examples of mis-classified instances from the Sports and Education domain. Examples are originally "Fake" but predicted as "Legitimate" .}
\label{tab:misclassified ins-1}
\end{table*}
\section{Conclusion and Future work}
In this article, we propose two deep learning based approaches to mitigate the problem of fake news detection from multiple domains platform. Antecedent works in this line pay attention on satirical news or made use of the content of the fact-checking websites, which was restricted to one domain (i.e. politics). To address these limitations, we focus to extend this problem into multi domain scenario. Our work extends the concept of fake news detection from uni-domain to multi-domain, thus making it more general and realistic. We evaluate our proposed models on the datasets whose contents are from multiple domains. Our two proposed approaches outperform the existing models with a notable margin. Experiments also reveal that there is a vital role of a domain in context of fake news detection. We would like to do more deeper analysis of the role of domain for this problem in future. Apart from this our future line of research would be as follows: 

\begin{itemize}
   \item It would be interesting for this work to encode the domain information in the Deep Neural Nets.
   \item BERT \cite{devlin-etal-2019-bert} and XLNet \cite{yang2019xlnet} embedding based model and make a comparison with fastText and ELMo based models in the context of fake news detection.
   \item Use of transfer learning and injection external knowledge for better understanding. 
   \item Handling of Named Entities efficiently and incorporate their embedding with the normal phrases. 
   \item Using WordNet to retrieve connections between words on the basis of semantics in the news corpora (both topic and document of news) which may influence in detection of Fake News.
\end{itemize}
\section{Acknowledgement}
This work is supported by \textbf{Elsevier Centre of Excellence for Natural Language Processing} at Computer Science and Engineering Department, Indian Institute of Technology Patna. Mr. Tanik Saikh acknowledges for the same.
\bibliography{acl2019}
\bibliographystyle{acl_natbib}

\appendix

\end{document}